\documentclass[letterpaper, 10 pt, journal, twoside]{IEEEtran} 
\usepackage{times}
\usepackage{etex} 
\usepackage{multicol}
\usepackage[bookmarks=true]{hyperref}
\usepackage{mathtools}
\usepackage{booktabs}
\usepackage{bm}
\usepackage{xcolor}
\usepackage[]{algorithm2e}
\usepackage{todonotes}
\SetKwInOut{Preprocessing}{Preprocessing}

\usepackage{pgfplots}
\pgfplotsset{compat=newest}
\usetikzlibrary{plotmarks}
\usetikzlibrary{babel}
\usetikzlibrary{arrows.meta}
\usepgfplotslibrary{patchplots}
\usepackage{grffile}
\usepackage{amsmath}

\usepackage{url}
\makeatletter
\g@addto@macro{\UrlBreaks}{\UrlOrds}
\makeatother

\usepackage[utf8]{inputenc}
\usepackage{pgfplots}
\pgfplotsset{compat=newest}
\usepgfplotslibrary{groupplots}
\usepgfplotslibrary{dateplot}
\RequirePackage{amsmath,amssymb,amsthm}
\RequirePackage{graphicx}

\newcommand{\bo}{\text{\boldmath$\omega$}}

\definecolor{Antonio}{rgb}{0.8,0.,0.8}

\definecolor{Mattia}{rgb}{0.,0.,0.8}

\DeclareMathOperator*{\argmin}{arg\,min}
\DeclareMathOperator{\EX}{\mathbb{E}}
\DeclareMathOperator{\V}{\mathbb{V}}

\newcommand\myeqone{\stackrel{\mathclap{\normalfont\mathrm{(1)}}}{=}}
\newcommand\myeqtwo{\stackrel{\mathclap{\normalfont\mathrm{(2)}}}{=}}
\newcommand\myeqthree{\stackrel{\mathclap{\normalfont\mathrm{(3)}}}{=}}
\newcommand\myeqfour{\stackrel{\mathclap{\normalfont\mathrm{(4)}}}{=}}
\newcommand\myapproxfive{\stackrel{\mathclap{\normalfont\mathrm{(5)}}}{\approx}}

\newcommand{\N}{\mathcal{N}}

\newcommand{\td}{\text{d}}

\newcommand{\x}{\mathbf{x}}

\newcommand{\sBb}{\mathtt{z}}

\newcommand{\y}{\mathbf{y}}

\newcommand{\z}{\mathbf{z}}

\newcommand{\X}{\mathbf{X}}
\newcommand{\Y}{\mathbf{Y}}

\newcommand{\I}{\mathbf{I}}

\newcommand{\Cov}{\text{Cov}}

\newcommand{\totvar}{\sigma_{tot}}
\newcommand{\sensnoise}{\mathbf{v}}
\newcommand{\sensnoisehat}{\mathbf{\hat{v}}}
\newcommand{\adfmean}{\pmb{\mu}}

\DeclarePairedDelimiterX{\infdivx}[2]{(}{)}{%
  #1\;\delimsize\|\;#2%
}
\newcommand{\kl}{\text{KL}\infdivx}
\DeclarePairedDelimiter{\norm}{\lVert}{\rVert}
\usepackage{nicefrac}
\newtheorem{theorem}{Theorem}[section]

\newtheorem{lemma}[theorem]{Lemma}

\definecolor{rebutall}{rgb}{0.8,0.,0.8}
\newcommand\rebutall[1]{\textcolor{black}{ #1}}

\ifCLASSINFOpdf
\else
\fi
\usepackage{array}

\ifCLASSOPTIONcompsoc
 \usepackage[caption=false,font=normalsize,labelfont=sf,textfont=sf]{subfig}
\else
 \usepackage[caption=false,font=footnotesize]{subfig}
\fi
\newcommand{\beginsupplement}{%
        \setcounter{table}{0}
        \renewcommand{\thetable}{S\arabic{table}}%
        \setcounter{figure}{0}
        \renewcommand{\thefigure}{S\arabic{figure}}%
     }

\begin{document}
%
\title{A General Framework \\ for Uncertainty Estimation in Deep Learning}

\author{Antonio Loquercio*$^{1}$, Mattia Segu*$^{1}$, and Davide Scaramuzza$^{1}$%
\thanks{Manuscript received: September, 10th, 2019; Revised November, 4th, 2019;
Accepted January, 14th, 2020.}
\thanks{This paper was recommended for publication by Editor Tamim Asfour upon
evaluation of the Associate Editor and Reviewers' comments. This work was supported by the Swiss National Center of
Competence Research Robotics (NCCR), through the Swiss National Science Foundation, and the
SNSF-ERC starting grant.}%
\thanks{$^{1}$All authors are with the Dep. of Informatics and Neuroinformatics of the University of Zurich and
ETH Zurich, Switzerland.}
\thanks{Digital Object Identifier (DOI): see top of this page.}
}


\markboth{IEEE Robotics and Automation Letters. Preprint Version. Accepted
January, 2020}
{Loquercio \MakeLowercase{\textit{et al.}}: A General Framework for Uncertainty Estimation}

\maketitle

\begin{abstract}
%
Neural networks predictions are unreliable when the input sample is out of the training distribution or corrupted by noise.
%
Being able to detect such failures automatically is fundamental to integrate deep learning algorithms into robotics.
Current approaches for uncertainty estimation of neural networks require changes to the network and optimization process, typically ignore prior knowledge about the data, and tend to make over-simplifying assumptions which underestimate uncertainty.
%
To address these limitations, we propose a novel framework for uncertainty estimation.
%
%
Based on Bayesian belief networks and Monte-Carlo sampling, our framework not only fully models the different sources of prediction uncertainty, but also incorporates prior data information, e.g.~sensor noise.
We show theoretically that this gives us the ability to capture uncertainty better than existing methods.
In addition, our framework has several desirable properties: (i) it is \emph{agnostic} to the network architecture and task; (ii) it does not require changes in the optimization process; (iii) it can be applied to \emph{already trained} architectures.
%
We thoroughly validate the proposed framework through extensive experiments on both computer vision and control tasks, where we outperform previous methods by up to 23\% in accuracy.
\end{abstract}

\section*{Supplementary Material}
%
%
The video available at \href{https://youtu.be/X7n-bRS5vSM}{https://youtu.be/X7n-bRS5vSM} shows qualitative results of our experiments.
The project's code is available at: \href{https://tinyurl.com/s3nygw7}{https://tinyurl.com/s3nygw7}
\let\thefootnote\relax\footnote{*These two authors contributed equally. Order is alphabetical.}

\section{Introduction}

\begin{figure}[t] \label{fig:first_img}
   \centering
   \includegraphics[width=\linewidth]{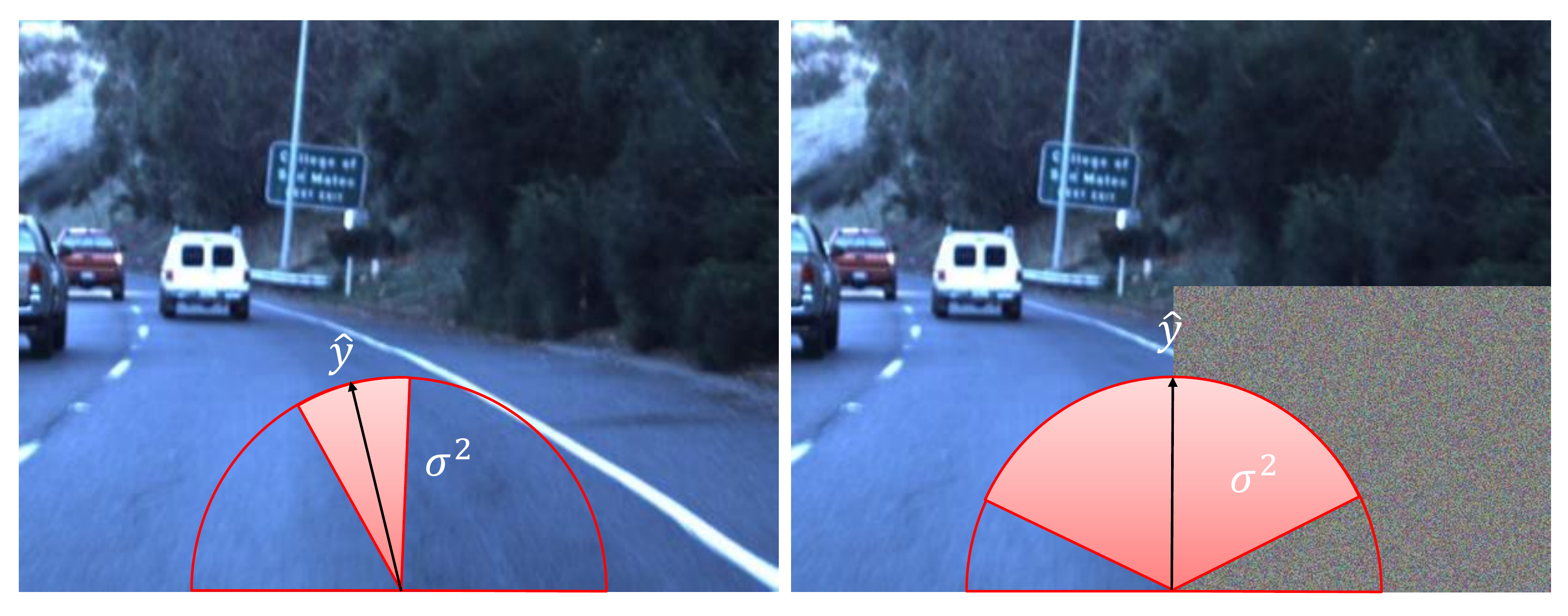}
   \caption{A neural network trained for steering angle prediction can be fully functional on a clean image (left) but generate unreliable predictions when processing a corrupted input (right).
   In this work, we propose a general framework to associate each network prediction with an uncertainty (illustrated above in red) that allows to detect such failure cases automatically.
   %
   %
}
   \label{img:first_img}
   \vspace{-3ex}
\end{figure}

\IEEEPARstart{R}{obots} act in an uncertain world.
In order to plan and make decisions, autonomous systems can only rely on noisy perceptions and approximated models.
Wrong decisions not only result in the failure of the mission but might even put human lives at risk, e.g., if the robot is an autonomous car (Fig.~\ref{fig:first_img}).
Under these conditions, deep learning algorithms can be fully integrated into robotic systems only if a measure of prediction uncertainty is available.
%
Indeed, estimating uncertainties enables Bayesian sensor fusion and provides valuable information during decision making~\cite{sunderhauf2018limits}.

Prediction uncertainty in deep neural networks generally derives from two sources: \emph{data} uncertainty and \emph{model} uncertainty.
The former arises because of noise in the data, usually caused by the sensors' imperfections.
The latter instead is generated from unbalances in the training data distribution.
For example, a rare sample should have higher model uncertainty than a sample which appears more often in the training data.
Both components of uncertainty play an important role in robotic applications.
A sensor can indeed never be assumed to be noise free, and training datasets cannot be expected to cover all the possible edge-cases.
%

%
Traditional approaches for uncertainty estimation model the network activations and weights by parametric probability distributions. 
%
%
However, these approaches are particularly difficult to train~\cite{hernandez2015probabilistic} and are rarely used in robotic applications.
Another family of approaches estimate uncertainties through sampling~\cite{Gal2015MCDO}.
%
Since they do not explicitly model data uncertainty, these methods generate over-confident predictions~\cite{osband2016risk}.
%
%
In addition, methods based on sampling generally disregard any relationship between data and model uncertainty, which increases the risk of underestimating uncertainties.
%
%
Indeed, an input sample with large noise should have larger model uncertainty than the same sample with lower noise.
%

%

In this paper, we propose a novel framework for uncertainty estimation of deep neural network predictions.
By combining Bayesian belief networks~\cite{Frey1999, Gast:2018:LPD, Boyen1998} with Monte-Carlo sampling, our framework captures prediction uncertainties better than state-of-the-art methodologies.
In order to do so, we propose two main innovations with respect to previous works: the use of prior information about the data, e.g., sensor noise, to compute data uncertainty, and the modelling of the relationship between data and model uncertainty.
We demonstrate both theoretically and experimentally that these two innovations allow our framework to produce higher quality uncertainty estimates than state-of-the-art methods.
%
%
%
In addition, our framework has some desirable properties: (i) it is \emph{agnostic} to the neural network architecture and task; (ii) it does not require any change in the optimization or learning process, and (iii) it can be applied to an \emph{already trained} neural network.
%
These properties make our approach an appealing solution to learning-based perception or control algorithms, enabling them to be better integrated into robotic systems.

To show the generality of our framework, we perform experiments on four challenging tasks: end-to-end steering angle prediction, obstacle future motion prediction, object recognition, and closed-loop control of a quadrotor.
%
%
In these tasks, we outperform existing methodologies for uncertainty estimation by up to 23\% in term of prediction accuracy.
However, our framework is not limited to these problems and can be applied, without any change, to a wide range of tasks.
Overall, our work makes the following contributions:
\begin{itemize}
    \item We propose a general framework to compute uncertainties of neural networks predictions. Our framework  is general in that it is agnostic to the network architecture, does not require changes in the learning or optimization process, and can be applied to already trained neural networks.
    \item We show mathematically that our framework can capture data and model uncertainty and use prior information about the data.
    \item We experimentally show that our approach outperforms existing methods for uncertainty estimation on a diverse set of tasks.
\end{itemize}{}

\section{Related Work}\label{sec:related_work}

%
%
%
In the following, we discuss the methods that have been proposed to estimate uncertainties and a series of approaches which have used this information in robotic systems.

\subsection{Estimating Uncertainties in Neural Networks Predictions}

A neural network is generally composed of a large number of parameters and non-linear activation functions, which makes the (multi-modal) posterior distribution of a network predictions intractable.
To approximate the posterior, existing methods deploy different techniques, mainly based on Bayesian inference and Monte-Carlo sampling.

To recover probabilistic predictions, Bayesian approaches represent neural networks weights through parametric distributions, e.g., exponential-family~\cite{ blundell2015weight, Frey1999, hernandez2015probabilistic, wang2016natural}.
Consequently, networks' predictions can also be represented by the same distributions, and can be analytically computed using non-linear belief networks~\cite{Frey1999} or graphical models~\cite{su2016nonlinear}.
More recently, Wang et al.~\cite{wang2016natural} propose natural parameter networks, which model inputs, parameters, nodes, and targets by Gaussian distributions.
%
%
Overall, these family of approaches can recover uncertainties in a principled way.
However, they generally increase the number of trainable parameters in a super-linear fashion, and require specific optimization techniques~\cite{hernandez2015probabilistic} which limits their impact in real-world applications.

In order to decrease the computational burden, Gast et al.~\cite{Gast:2018:LPD} proposed to replace the network's input, activations, and outputs by distributions, while keeping network's weights deterministic.
\rebutall{Similarly, probabilistic deep state space models retrieve data uncertainty in sequential data, and use it for learning-based filtering~\cite{fraccaro2017disentangled, wu2018deep}.
However, disregarding weights uncertainty generally results in over-confident predictions, in particular for inputs not well represented in the training data.}

Instead of representing neural networks parameters and activations by probability distributions, a second class of methods proposed to use Monte-Carlo (MC) sampling to estimate uncertainty.
The MC samples are generally computed using an ensemble of neural networks.
The prediction ensemble could either be generated by differently trained networks~\cite{Lee2018, kahn2018self, lakshminarayanan2017simple}, or by keeping drop-out at test-time~\cite{Gal2015MCDO}.
While this class of approaches can represent well the multi-modal posterior by sampling, it cannot generally represent data uncertainty, due for example to sensor noise.
A possible solution is to tune the dropout rates~\cite{gal2017concrete}, however it is always possible to construct examples where this approach would generate erroneous predictions~\cite{osband2016risk}.

To model data uncertainty, Kendall et al.~\cite{Kendall2017} proposed to add to each output a ``variance" variable, which is trained by a maximum-likelihood (a.k.a. heteroscedastic) loss on data.
Combined with Monte-Carlo sampling, this approach can predict both the model and data uncertainty. 
However, this method requires to change the architecture, due to the variance addition, and to use the heteroscedastic loss for training, which is not always feasible.

Akin to many of the aforementioned methods, we use Monte-Carlo samples to predict model uncertainty.
Through several experiments, we show why this type of uncertainty, generally ignored or loosely modelled by Bayesian methods~\cite{Gast:2018:LPD}, cannot be disregarded.
In addition to Monte-Carlo sampling, our approach also computes the prediction uncertainty due to the sensors noise by using gaussian belief networks~\cite{Frey1999} and assumed density filtering~\cite{Boyen1998}.
Therefore, our approach can recover the full prediction uncertainty for any given (and possible already trained) neural network, without requiring any architectural or optimization change.

\subsection{Uncertainty Estimation in Robotics}
  
Given the paramount importance of safety, autonomous driving research has allocated a lot of attention to the problem of uncertainty estimation, from both the perception~\cite{Feng2018, neumann2018relaxed} and the control side~\cite{Lee2018, kahn2018self}.
%
%
Feng. et al.~\cite{Feng2018} showed an increase in performance and reliability of a 3D Lidar vehicle detection system by adding uncertainty estimates to the detection pipeline.
Predicting uncertainty was also shown to be fundamental to cope with sensor failures in autonomous driving~\cite{Lee2018},
and to speed-up the reinforcement learning process on a real robot~\cite{kahn2018self}.

For the task of autonomous drone racing, Kaufmann et al.~\cite{kaufmann2018beauty} demonstrated the possibility to combine optimal control methods to a network-based perception system by using uncertainty estimation and filtering.
Also for the task of robot manipulation, uncertainty estimation was shown to play a fundamental role to increase the learning efficiency and guarantee the manipulator safety~\cite{stulp2011learning, chua2018deep}.

In order to fully integrate deep learning into robotics, learning systems should reliably estimate the uncertainty in their predictions~\cite{sunderhauf2018limits}. 
Our framework represents a minimally invasive solution to this problem:~we do not require any architectural changes or re-training of existing models.

\begin{figure*}[t]
    \centering
    \includegraphics[width=0.9\textwidth]{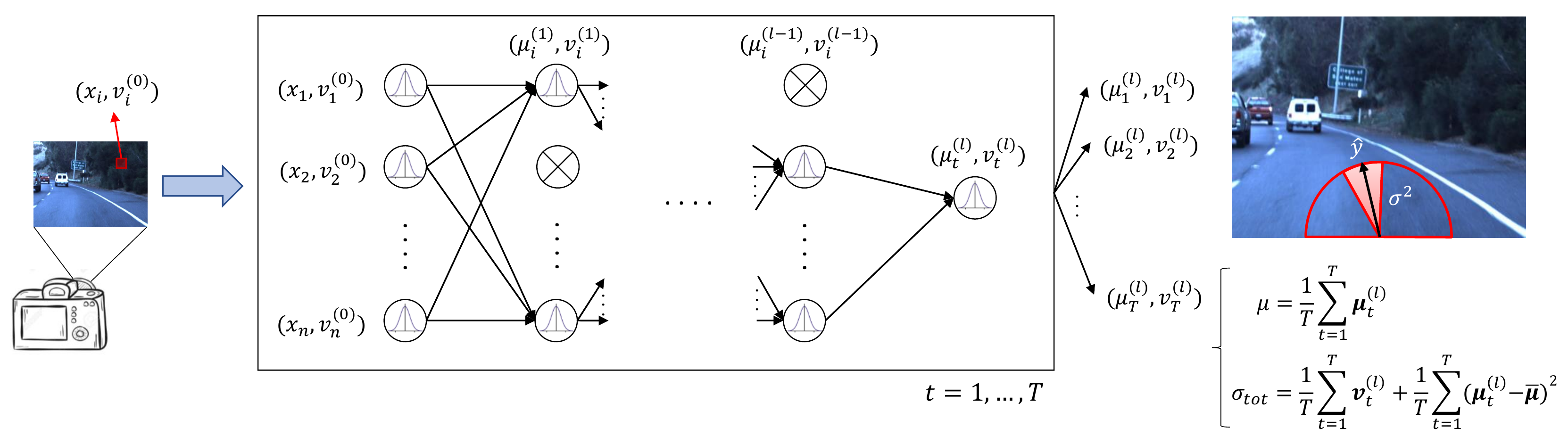}
    \caption{Given an input sample $\x$, associated with noise $\sensnoise^{(0)}$, and a trained neural network, our framework computes the confidence associated to the network output. In order to do so, it first transforms the given network into a Bayesian belief network. Then, it uses an ensemble of $T$ such networks, created by enabling dropout at test time, to generate the final prediction $\adfmean$ and uncertainty $\totvar$.}
    \label{fig:method_summary}
    \vspace{-3mm}
\end{figure*}{}

\section{Methodology}\label{sec:methodology}
Due to the large number of (possibly non-linear) operations required to generate predictions, the posterior distribution  $p(\y|\x)$, where $\y$ are output predictions and $\x$ input samples, is intractable.
Formally, we define the total prediction uncertainty as $\totvar=\text{Var}_{p(\y|\x)}(\y)$.
This uncertainty comes from two sources: data and model uncertainty.
In order to estimate $\totvar$, we derive a tractable approximation of $p(\y|\x)$.
%
In the following, we present the derivation of this approximation by using Bayesian inference, and the resulting algorithm to predict $\totvar$ (illustrated in Fig.~\ref{fig:method_summary}).

\subsection{The data uncertainty}\label{sec:data_uncertainty}

\noindent Sensors' observations, e.g. images, are generally corrupted by noise.
Therefore, what a neural network processes as input is $\z$, a noisy version of the ``real" input $\x$.
We assume that the sensor has known noise characteristic $\sensnoise$, which can be generally acquired by system identification or hardware specifications.
Given $\sensnoise$, we assume the input data distribution $q(\z|\x)$ to be:
\begin{equation}
    q(\z|\x) \sim \N \left( \z; \x, \sensnoise \right).
\end{equation}
\noindent The output uncertainty resulting from this noise is generally defined as \emph{data} (or aleatoric) uncertainty.

In order to compute data uncertainty, we  forward-propagate sensor noise through the network via Assumed Density Filtering (ADF)~\cite{Boyen1998}.
This approach, initially applied to neural networks by Gast et al.~\cite{Gast:2018:LPD}, replaces each network activation, including input and output, by probability distributions.
Specifically, the joint density of all activations in a network with $l$ layers is:
\begin{align}
    p(\z^{(0:l)}) &= p(\z^{(0)})\prod_{i=1}^{l}p(\z^{(i)}|\z^{(i-1)}) \\
    p(\z^{(i)}|\z^{(i-1)}) &= \delta[\z^{(i)} - \mathbf{f}^{(i)}(\z^{(i-1)})]
\end{align}{}
\noindent where $\delta[\cdot]$ is the Dirac delta and $\mathbf{f}^{(i)}$ the i-th network layer.
Since this distribution is intractable, ADF approximates it with:
\begin{equation}
    p(\z^{(0:l)}) \approx q(\z^{(0:l)}) = q(\z^{(0)}) \prod_{i=1}^{l}q(\z^{(i)})
\end{equation}
where \rebutall{$q(\z)$} is normally distributed, with all components independent:
\rebutall{
\begin{equation}
    q(\z^{(i)}) \sim \N \left( \z^{(i)}; \adfmean^{(i)}, \sensnoise^{(i)} \right) = \prod_j \N \left( \z^{(i)}_{j}; \adfmean^{(i)}_{j}, \sensnoise^{(i)}_{j} \right).
\end{equation}
}
The activation $\z^{(i-1)}$ is then processed by the (possibly non-linear) i-th layer function, $\mathbf{f}^{(i)}$ , which transforms it into the (not necessarily normal) distribution:
\begin{equation}
    \hat{p}(\z^{(0:i)}) = p(\z^{(i)}|\z^{(i-1)}) q(\z^{(0:i-1)}).
\end{equation}
\noindent The goal of ADF is then to find the distribution $ q(\z^{(0:i)})$ which better approximates $\hat{p}(\z^{(0:i)})$ under some measure, e.g. Kullback-Leibler divergence:
\begin{equation}
    q(\z^{(0:i)}) = \argmin_{\hat{q}(\z^{(0:i)})} \kl{\hat{q}(\z^{(0:i)})}{\hat{p}(\z^{(0:i)})}
    \label{eq:kl_div}
\end{equation}{}
Minka et al.~\cite{minka2001family} showed that the solution to \eqref{eq:kl_div} requires matching the moments of the two distributions. Under the normality assumptions, this is equivalent to:
\begin{align}
    \adfmean^{(i)} &= \EX_{q(\z^{(i-1)})} [\ \mathbf{f}^{(i)}(z^{(i-1)}) ]\ \label{eq:adf_mean} \\
    \sensnoise^{(i)} &= \V_{q(\z^{(i-1)})} [\ \mathbf{f}^{(i)}(z^{(i-1)}) \label{eq:adf_var} ]\
\end{align}{}
\noindent where $\EX$ and $\V$ are the first and second moment of the distribution.
The solution of Eq.~\eqref{eq:adf_mean} and Eq.~\eqref{eq:adf_var} can be computed analytically for the majority of functions used in neural networks, e.g. convolution, de-convolutions, relu, etc, \rebutall{and has an approximated solution for max-pooling.}
This results in a recursive formula to compute the activations mean and uncertainty, $(\adfmean^{(i)}, \sensnoise^{(i)})$, given the parameters of the previous activations distribution $q(\z^{(i-1)})$.
We refer the reader to ~\cite{Gast:2018:LPD, Frey1999, Jin2016} for the details of the propagation formulas.

In summary, ADF modifies the forward pass of a neural network to generate not only output predictions $\adfmean^{(l)}$, but also their respective data uncertainties $\sensnoise^{(l)}$.
In order to do so, ADF propagates the input uncertainty $\sensnoise = \sensnoise^{(0)}$, which, in a robotics scenario, corresponds to the sensor noise characteristics.

\subsection{The model uncertainty}\label{sec:model_uncertainty}

Model (or epistemic) uncertainty refers to the confidence a model has about its prediction.
%
%
Similarly to Bayesian approaches~\cite{Denker1991, MacKay1992, Neal1995, Srivastava2014dropout, Gal2015MCDO}, we represent this uncertainty by placing a distribution over the neural network weights, $\bo$.
This distribution depends on the training dataset $\text{D} = \{\X,\Y\}$, where $\X,\Y$ are training samples and labels, respectively.
Therefore, the weight distribution after training can be written as $p(\bo|\X,\Y)$.

Except in trivial cases, $p(\bo|\X,\Y)$ is intractable.
In order to approximate this distribution, Monte-Carlo based approaches collect weights samples by using dropout at test time~\cite{Srivastava2014dropout, Gal2015MCDO, Kendall2017}.
Formally, this entails to approximate:
\begin{equation}
    p(\bo|\X,\Y) \approx q(\bo;\Phi) = Bern(\bo; \Phi)
\end{equation}
\noindent where $\Phi$ are the Bernoulli (or dropout) rates on the weights.
Under this assumption, the model uncertainty is the variance of $T$ Monte-Carlo samples, i.e.~\cite{Gal2015MCDO}:
\begin{equation}
    \text{Var}_{p(\y|\x)}^{model}(\y) = \sigma_{model} = \frac{1}{T} \sum_{t=1}^T (\y_t - \bar{\y})^2
    \label{eq:model_var}
\end{equation}
\noindent where $\{\y_t\}_{t=1}^T$ is a set of $T$ sampled outputs for weights instances $\bo^t \sim q(\bo; \Phi)$ and $\bar{\y} = \nicefrac{1}{T}\sum_{t} \y_t$.

Eq.~\ref{eq:model_var} has an intuitive explanation: Due to over-parametrization, a network develops redundant representations of samples frequently observed in the training data.
Because of the redundancy, predictions for those samples will remain approximately constant when a part of the network is switched off with dropout. Consequently, these samples will receive a low model uncertainty. 
In contrast, for rare samples the network is not able to generate such redundancies. Therefore, it will associate them with high model uncertainty.

\subsection{Model uncertainty of an already trained network}
The optimal dropout rates $\Phi$ needed to compute $\sigma_{model}$ are the minimizers of the distance between the real and the hypothesis weight distribution:
\begin{equation}
    \Phi = \argmin_{\hat{\Phi}} \kl{p(\bo|\X,\Y)}{q(\bo;\hat{\Phi})}.
    \label{eq:optimal_Phi}
\end{equation}
Previous works showed that the best $\Phi$ corresponds to the training dropout rates~\cite{Srivastava2014dropout, Gal2015MCDO, Kendall2017}.
This, however, hinders the computation of model uncertainty for networks trained without dropout.

Since re-training a given network with a specific rate $\Phi$ is not always possible for several applications, we propose to find the best $\Phi$ \emph{after training} by minimizing the negative log-likelihood between predicted and ground-truth labels.
This is justified by the following lemma:
\begin{lemma}
The dropout rates $\Phi$ minimizing Eq.~\eqref{eq:optimal_Phi}, under normality assumption on the output, are equivalent to:
\begin{equation}
    \Phi = \argmin_{\hat{\Phi}} \sum_{d \in D} \frac{1}{2}\log(\totvar^{d}) + \frac{1}{2\totvar^{d}}(\y_{gt}^{d}-\y_{pred}^{d}(\hat{\Phi)})^2.
    \label{eq:compute_phi}
\end{equation}
\end{lemma}{}
%
\begin{proof}
A successfully trained network $p_{net}(\y|\x,\bo)$ can very well predict the ground-truth, \emph{i.e.}:
\begin{equation}
    p_{gt}(\y|\x) \approx p_{pred}(\y|\x) = \int_{\bo}{p_{net}(\y|\x,\bo) p(\bo|\X,\Y)}.
    \label{eq:gt_dist}
\end{equation}{}
By approximating $p(\bo|\X,\Y)$ by $q(\bo|\Phi)$, \emph{i.e.}, putting dropout only at test time, the real predicted distribution is actually:
\begin{equation}
   \hat{p}_{pred}(\y|\x; \Phi) = \int_{\bo}{p_{net}(\y|\x,\bo) q(\bo|\Phi)}.
   \label{eq:pred_dist}
\end{equation}
Since $p_{net}(\cdot)$ in Eq.~\eqref{eq:gt_dist} and Eq.~\eqref{eq:pred_dist} are the same, minimizing $\kl{p_{gt}(\y|\x)}{\hat{p}_{pred}(\y|\x; \Phi)}$  is equivalent to minimizing Eq.~\eqref{eq:optimal_Phi}.
Assuming that both $p_{net}$ and $p_{gt}$ are normal, and that $\sigma_{gt} \to 0$ (i.e. ground-truth is quasi-deterministic), the distance between the predicted and ground-truth distribution is equivalent to Eq.~\eqref{eq:compute_phi}.
\end{proof}
\noindent Practically, $\Phi$ is found by grid-search on a log-range of 20 possible rates in the range $[0,1]$.


\subsection{The total uncertainty}

Section~\ref{sec:data_uncertainty} shows that ADF can be used to propagate sensor uncertainties to the network outputs.
This is equivalent to model the output distribution $p(\y|\x) \approx p(\y|\z,\bo)p(\z|\x)$, where $\bo$ are deterministic network parameters and $p(\z|\x)$ the sensor noise characteristics.
Instead, Section~\ref{sec:model_uncertainty} shows that model uncertainty can be computed by putting a distribution on the network weights $p(\bo|\X,\Y)$.
The total uncertainty $\totvar$ results from the combination of the model and data uncertainty.
%
It can be computed through a stochastic version of ADF, as presented in the following lemma.
\begin{lemma}\label{lemma:totvar}
The total variance of a network output $\y$ for an input sample $\x$ corrupted by noise $\sensnoise^{(0)}$ is:
\begin{equation} \label{eq:tot_var}
\totvar = \text{Var}_{p(\y|\x)}(\y)= \frac{1}{T} \sum_{t=1}^T \sensnoise^{(l)}_{t} + (\adfmean_t^{(l)} - \bar{\adfmean})^2
\end{equation}
\noindent where $\{\adfmean_t^{(l)},\sensnoise^{(l)}_{t}\}_{t=1}^T$ is a set of $T$ outputs from the ADF network with weights $\bo^t \sim q(\bo; \Phi)$ and $\bar{\adfmean} = \nicefrac{1}{T}\sum_{t} \adfmean_t^{l}$.
\end{lemma}
\noindent Its proof can be found in the supplementary material.
Intuitively, Eq.~\eqref{eq:tot_var} generates the total uncertainty by summing the two components of data and model uncertainty.
Note the difference between Eq.~\eqref{eq:model_var} and our model uncertainty, $\nicefrac{1}{T}\sum_{t=1}^T(\adfmean_t^{(l)} - \bar{\adfmean})^2$.
Differently from Eq.~\eqref{eq:model_var}, the prediction ensemble used to calculate the model variance is not generated with network outputs $\y_t$, but with ADF predictions $\adfmean^{(l)}_t$.
Consequently, the model uncertainty also depends on the input sensor noise $\sensnoise^{(0)}$.
Indeed, this is a very intuitive result: even though a sample has been frequently observed in the training data, it should have large model uncertainty if corrupted by high noise.
From Lemma~\ref{lemma:totvar} we derive a computationally feasible algorithm to compute, at the same time, predictions and total uncertainties.
Illustrated in Fig.~\ref{fig:method_summary}, the algorithm is composed of three main steps: (i) transforming a neural network into its ADF version (which does not require re-training), (ii) collect T samples by forwarding $(\x,\sensnoise^{(0)})$ to the network with $\bo^t \sim q(\bo; \Phi)$ and (iii) compute output predictions and variances according to lemma~\ref{lemma:totvar}.

 
 

It is interesting to draw a connection between Eq.~\eqref{eq:tot_var} and the total uncertainty formulas used by previous works.
Gast et al.~\cite{Gal2015MCDO}, for example, do not use ADF networks to collect Monte-Carlo samples, and substitutes to the data uncertainty $\sensnoise^{(l)}$ a user-defined constant $\sigma_d$.
Using a constant for the data uncertainty is nevertheless problematic, since different input samples might have different noise levels (due to, for example, temperature or pressure changes).
Manually tuning this constant is generally difficult in practice, since it is not possible to use prior information about sensor noise characteristics.
This makes it less attractive to robotics applications, where this information is either available or can be retrieved via identification.

In order to increase adaptation, Kendall et al.~\cite{Kendall2017} proposed to learn the data uncertainty from the data itself.
However, this comes at the cost of modifying the architecture and the training process, which hinders its application to already trained models and generally results in performance drops.
Moreover, it considers the model and data uncertainty to be completely independent, which is a overly-restrictive assumption in many cases.
For example, high sensor noise can result in large model uncertainty, in particular if the model was never exposed, at training time, to such kind of noise levels.
In contrast, our approach can model this dependence, since it uses ADF samples to compute model uncertainty.
We refer the reader to the proof of Lemma~\ref{lemma:totvar} for the formal justification of the previous statements.

\section{Experiments}\label{chap:experiments}

We validate our framework to compute uncertainties on several computer vision and robotic tasks.
Specifically, as demonstrators we select end-to-end steering angle prediction, object future motion prediction, object recognition, and model-error compensation for autonomous drone flight.
These demonstrators encompass the most active areas of research in mobile robotics, from computer vision to learning-based control.
For each application, we compare against state-of-the-art methods for uncertainty estimation both qualitatively and quantitatively.
\rebutall{All training details are reported in the appendix.}
%

\subsection{Demonstrators}

\begin{figure}
\centering
\renewcommand{\arraystretch}{0}
\begin{tabular}{c c}
    \includegraphics[width=0.47\linewidth]{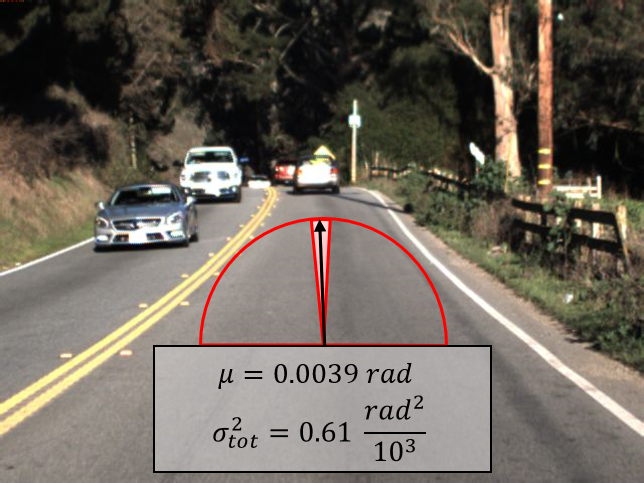} &
    \includegraphics[width=0.47\linewidth]{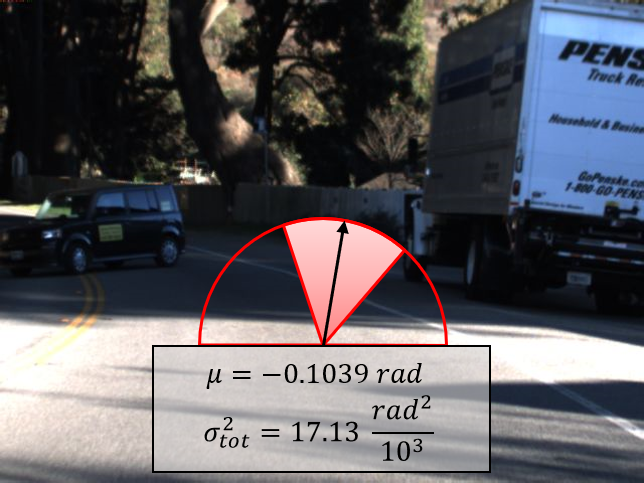} \\
\end{tabular}
    \caption{On well illuminated frames where the car is driving straight, a network trained to predict steering angles is very certain about its output (left).
    In contrast, for poorly illuminated, ambiguous frames, the network is highly uncertain about its predictions (right).}
    \label{fig:steering_angles_var}
    \vspace{-1mm}
\end{figure}

\noindent \textbf{End-to-End Steering Angle Prediction:}
\begin{table}[t]
\centering
\begin{tabular}{l c c c c } 
 \toprule
 Method & Re-train & RMSE & EVA & NLL \\
 \midrule
 Gal et al.~\cite{Gal2015MCDO} & Yes & 0.09 & \textbf{0.83} & -0.72 \\ 
 Gast et al.~\cite{Gast:2018:LPD} & Yes & \rebutall{0.10} & \rebutall{0.79} & \rebutall{-0.89} \\
 Kendall et al.~\cite{Kendall2017} & Yes & 0.11 & 0.75 & \textbf{-1.1} \\
 Ours & No & \textbf{0.09} & 0.81 & -1.0 \\
 \bottomrule \\
\end{tabular}
\caption{\small Benchmark comparison against state-of-the-art methods for variance estimation on the end-to-end steering angle prediction task.}
\vspace{-8mm}
\label{table:quantitative_comparison}
\end{table}
Neural networks trained to predict steering angles from images have been used in several robotics applications, e.g. autonomous driving~\cite{bojarski2016end, muller2018driving} and flying~\cite{Loquercio_2018}.
In order to account for the safety of the platform, however, previous works showed the importance of quantifying uncertainties~\cite{kahn2018self, Lee2018}.
In this section, we show that our framework can be used to estimate uncertainty without losing prediction performance on steering prediction.

To predict steering angles, we use the DroNet architecture of Loquercio et al.~\cite{Loquercio_2018}, since it was shown to allow closed-loop control on several unmanned aerial platforms~\cite{palossi201964mw, kaufmann2018deep}.
Differently from DroNet, however, we add a dropout layer after each convolution or fully connected layer. This is indeed necessary to extract Monte-Carlo samples.
In order to show that our approach can estimate uncertainties from already trained networks, dropout is only activated \emph{at test time}.
We train this architecture on the Udacity dataset~\cite{Loquercio_2018}, which provides labelled images collected from a car in a large set of environments, illumination and traffic conditions.
As it is the standard for the problem~\cite{bojarski2016end, Loquercio_2018}, we train the network with mean-squared-error loss $\norm{y_{gt} - y_{pred}}^2$, where $y_{gt}$ and $y_{pred}$ are the ground-truth and estimated steerings.
For evaluation, we measure performance with Explained Variance (EVA) and Root Mean Squared Error (RMSE), as in~\cite{Loquercio_2018}.
Since there is no ground-truth for variances, we quantify their accuracy with negative log-likelihood (NLL) $\nicefrac{1}{2}\log(\sigma_{tot}) + \nicefrac{1}{2\sigma_{tot}}(y_{gt}-y_{pred})^2$~\cite{Gal2015MCDO, Kendall2017, Gast:2018:LPD}.

We compare our approach against state-of-the-art methods for uncertainty estimation. For a fair comparison, all methods share the same network architecture.
For all the methods using sampling, we keep the number of samples fixed to $T=20$, as it allows real-time performance (see Sec~\ref{sec:practical}).
Our approach additionally assumes an input noise variance $\sensnoise=2$ grayscale levels, which is typical for the type of camera used in the Udacity dataset.

Table~\ref{table:quantitative_comparison} summarizes the results of this experiment.
Unsurprisingly, the method of Kendall et al.~\cite{Kendall2017}, trained to minimize the NLL loss, can predict good variances, but loses prediction quality due to the change of training loss and architecture.
\rebutall{The approach of Gast et al.~\cite{Gast:2018:LPD}, trained under the same NLL objective, performs analogously in terms of RMSE and EVA. 
However, it performs worse in term of NLL, since this baseline only accounts for data uncertainty.
In contrast to the previous baselines, the method of Gal et al.~\cite{Gal2015MCDO} predicts more precise steering angles due to ensembling, but generates poorer uncertainty estimates.}
With our framework, it is not necessary to make compromises: we can both make accurate predictions and have high quality uncertainty estimates, without changing or re-training the network.

Fig.~\ref{fig:steering_angles_var} shows some qualitative results of our approach.
As expected, our approach assigns very low variance to well-illuminated images with $y_{gt} \approx 0$.
These are indeed the most frequent samples in the training dataset, and contain limited image noise.
In contrast, our method predicts high uncertainties for images with large light gradients.
This is expected, since those samples can be ambiguous and have a smaller signal to noise ratio.
For more qualitative experiments, we refer the reader to the supplementary video.

\noindent\textbf{Object Future Motion Prediction:} In this section, we train a neural network to predict the motion of an object in the future.
Endowing robots with such an ability is important for several navigation tasks, e.g. path planning and obstacle avoidance.
More specifically, the task is equivalent to predict the position of an object at time $t + \Delta t$, assuming to have the video of the object moving from $t=0,\dots,t$.
For the sake of simplicity, we predict motion only in the image plane, or, equivalently, the future 2D optical flow of the object.
%

\begin{figure}[t]
\centering
\renewcommand{\arraystretch}{0}
\begin{tabular}{c}
    \includegraphics[width=0.8\linewidth]{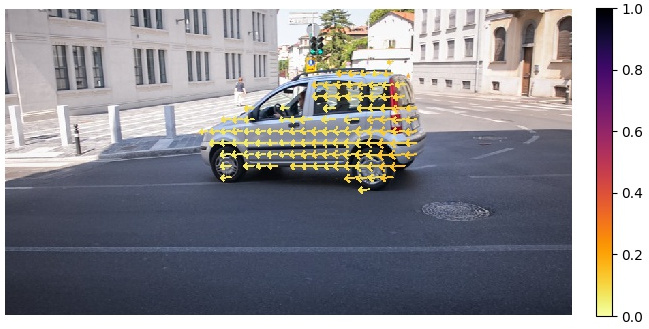} \\
    \includegraphics[width=0.8\linewidth]{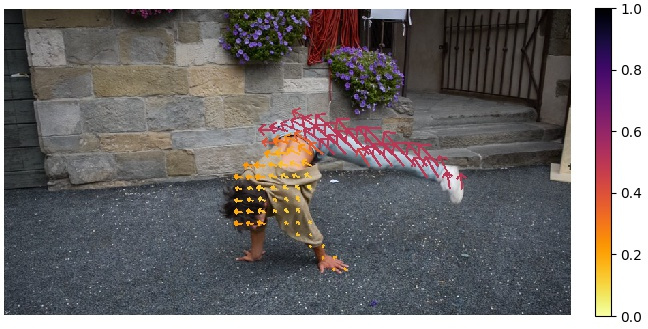} \\
\end{tabular}
    \caption{Qualitative evaluation on the task of object future motion prediction. Future motion predictions are indicated by arrows, while uncertainties are color-coded. For objects whose motion is easily predictable, e.g car on the top, our framework produces good predictions with low uncertainties. In contrast, for object whose motion is not easily predictable, e.g. the dance at the bottom, predictions are associated with higher variance.}
    \label{fig:object_motion_var}
    \vspace{-4mm}
\end{figure}

In order to predict future flows we use the Flownet2S architecture~\cite{ilg2017flownet}, as it represents a good trade-off between performance and computational cost.
The input to the network consists of a concatenation of the current and past frames $I_{t}, I_{t-1}$, and the object mask $M_{t}$.
Its output is instead the optical flow \emph{on the object} between the current and the (unobserved) future frame $I_{t}, I_{t+1}$.
Ground-truth for this task is generated by the Flownet2 architecture~\cite{ilg2017flownet}, which is instead provided with the current and future frames $I_{t}, I_{t+1}$.

We perform experiments on the Davis 2016 dataset~\cite{Perazzi_2016}, which has high quality videos with moving objects, as well as  pixel-level object masks annotations.
Also for this experiment, it is assumed an input noise variance $\sensnoise^{(0)}$ of 2 pixels, which is compatible with the type of camera used to collect the dataset.

We again compare our framework for uncertainty estimation to state-of-the-art approaches~\cite{Gal2015MCDO, Gast:2018:LPD, Kendall2017}.
To quantitatively evaluate the optical flow predictions, we use the standard end-point error (EPE) metric~\cite{ilg2017flownet}.
This metric is, however, `local', since it does not evaluate the motion prediction \emph{as a whole} but just as average over pixels.
In order to better understand if our approach can predict the motion of the entire object correctly, we fit a Gaussian to both the sets of predicted and ground-truth flows.
The KL distance between these two distributions represents our second metric.
Finally, we use the standard negative log-likelihood metric (NLL) to evaluate the uncertainty estimations.

Table~\ref{table:quantitative_comparison_obj} summarizes the results of this experiment.
Our method outperforms all baselines on every metric.
Interestingly, even though our network has not been specifically trained to predict variances as in Kendall et al.~\cite{Kendall2017},
it estimates uncertainty 23\% better than the runner-up method.
At the same time, being the network specifically trained for the task, it makes accurate predictions, outperforming the approach from Gal et al.~\cite{Gal2015MCDO} by 2\% in terms of RMSE and 20\% on the KL metric.
%

\begin{table}[t]
\centering
\begin{tabular}{l c c c c } 
 \toprule
 Method & Re-train & EPE & KL & NLL \\
 \midrule
 Gal et al.~\cite{Gal2015MCDO} & Yes & 5.99 & 56.7  & 6.96 \\ 
 Gast et al.~\cite{Gast:2018:LPD} & Yes & \rebutall{6.12} & \rebutall{50.1} & \rebutall{5.74} \\
 Kendall et al.~\cite{Kendall2017} & Yes & 6.79 & 52.5 & 5.28 \\
 Ours & No & \textbf{5.91} & \textbf{45.1} & \textbf{4.07} \\
 \bottomrule \\
\end{tabular}
\caption{\small Benchmark comparison against state-of-the-art on the task of object future motion prediction.}
\vspace{-7mm}
\label{table:quantitative_comparison_obj}
\end{table}

\begin{figure*}[t]
\centering
\renewcommand{\arraystretch}{0}
\begin{tabular}{c c c}
    \includegraphics[width=0.31\textwidth]{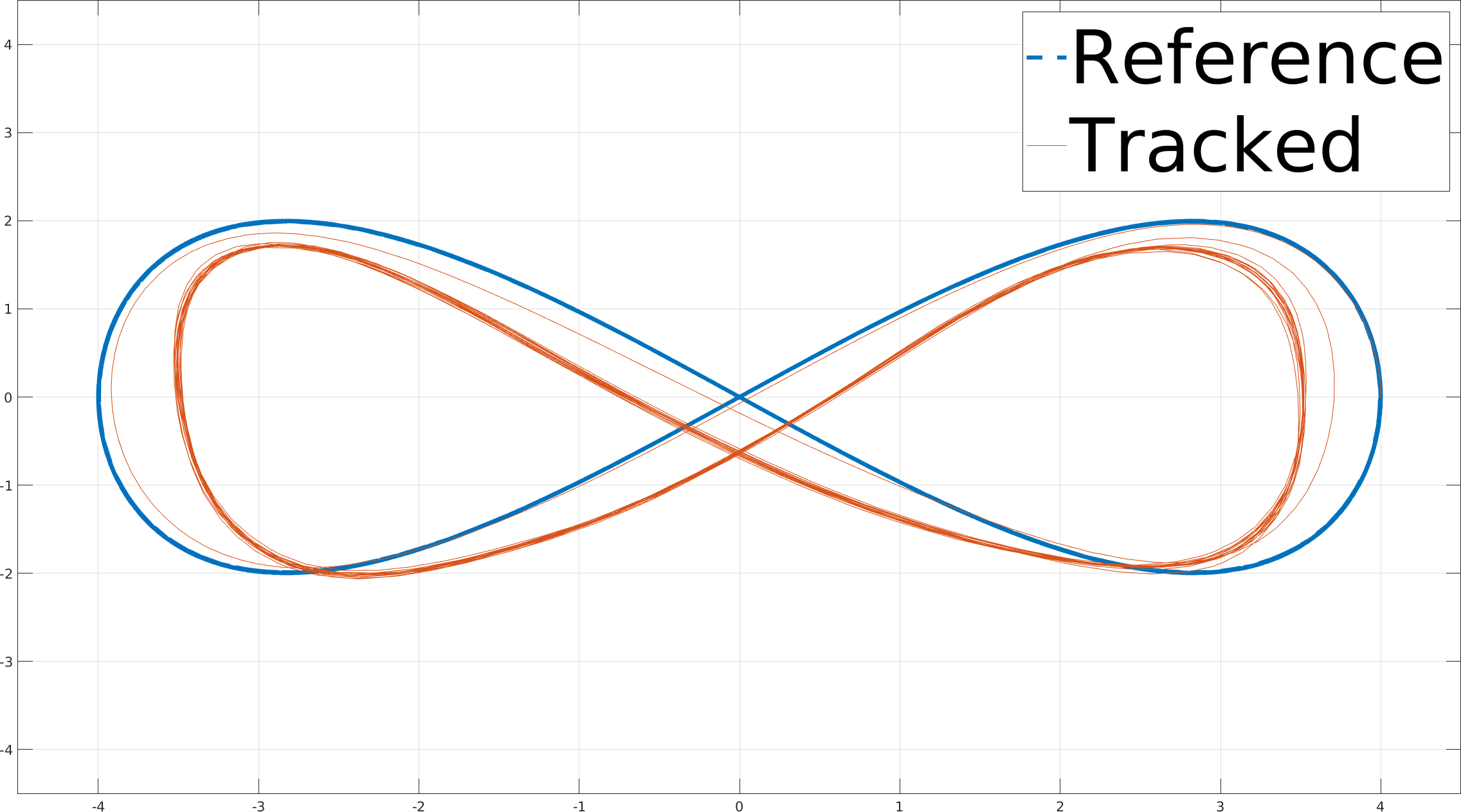} &
    \includegraphics[width=0.31\textwidth]{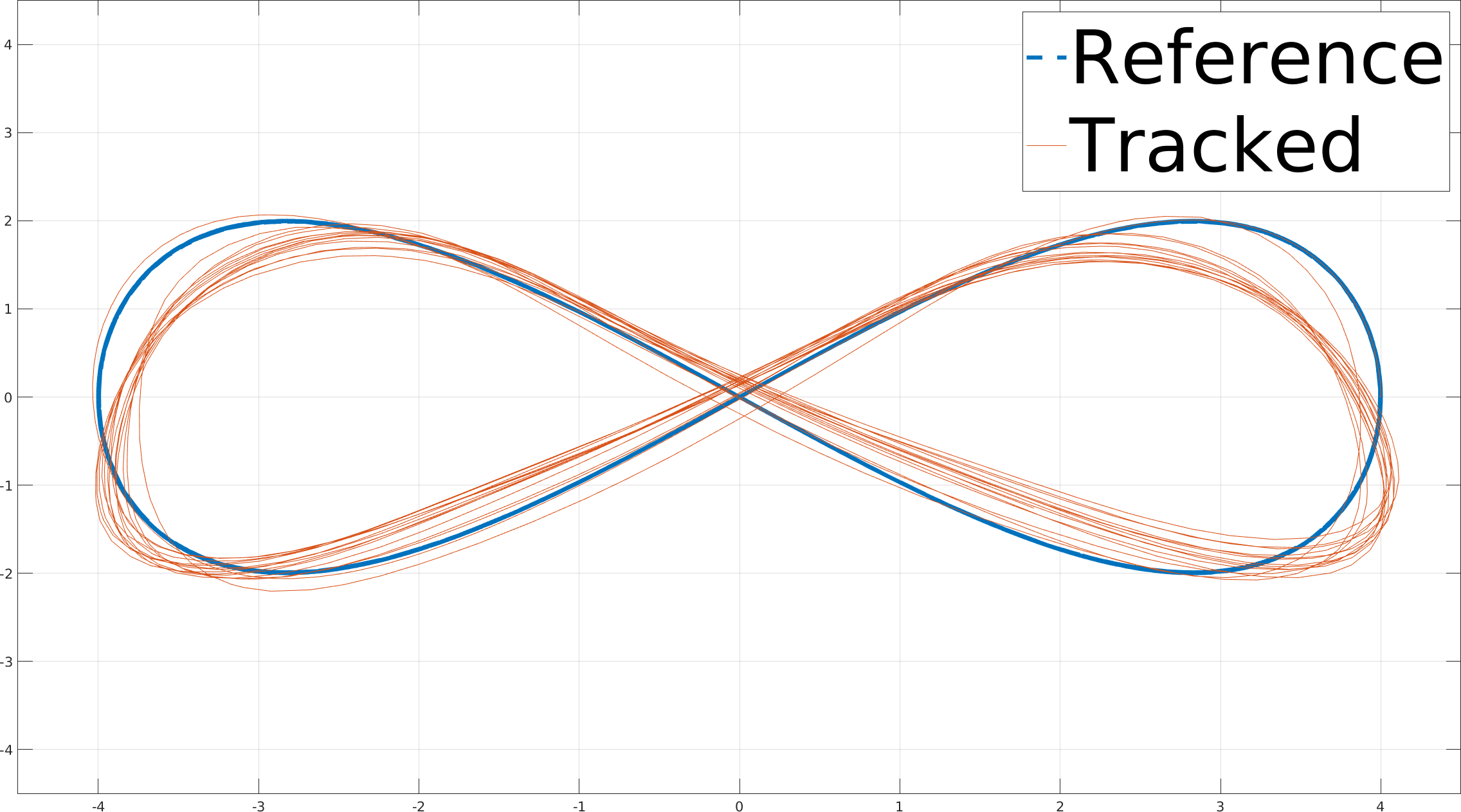} & \includegraphics[width=0.31\textwidth]{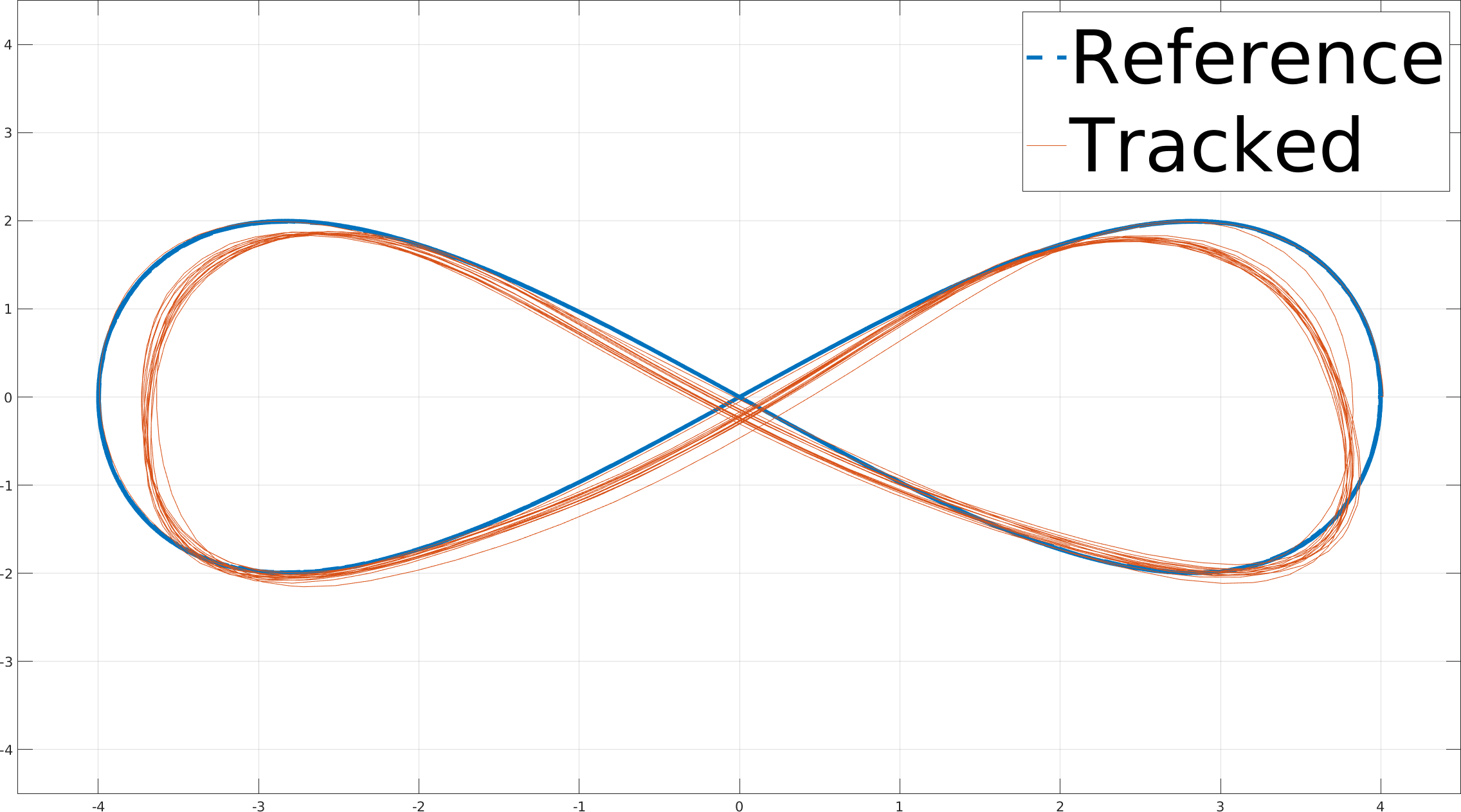} \\
\end{tabular}
    \caption{Qualitative comparison between different control models in a lemniscate trajectory. Due to drag effects, the nominal model (left) cannot accurately track the reference trajectory. Linear model compensation~\cite{faessler2018differential} (middle) decreases tracking errors, but still provides suboptimal results.
    Our approach (right) providing non-linear compensations to the model only if the prediction uncertainty is within a confidence bound, achieves the best tracking performance.}
    \label{fig:closed_loop}
    \vspace{-3mm}
\end{figure*}

Qualitative results in Fig.~\ref{fig:object_motion_var} show that our framework captures an intuitive behaviour of the prediction network.
Whenever the motion of the object is highly predictable, e.g. a car driving on the road, future optical flow vectors are accurately estimated and a low uncertainty is assigned to them.
In contrast, if the object moves unpredictably, e.g. a dancer, the network is more uncertain about its predictions, particularly for the parts of the person which quickly change velocity. 

\noindent \textbf{Object Recognition:} In  this  section,  we  investigate  the performance of our framework on a classic computer vision task: object recognition.
In order to do that, we evaluate our framework on the CIFAR-10 Dataset.
We use two metrics to evaluate the performance of our approach: the average classification accuracy, and the average of per-class negative log-likelihood.
%
%
Results of this evaluation are reported in Table~\ref{tab:cifar_results}.
Similarly to previous tasks, variance estimation in object recognition benefits from considering both model and data uncertainty.
%
%
The aforementioned result table does not include the baseline of Kendall et al.~\cite{Kendall2017}, since its training procedure is specifically designed for regression problems, and it failed to converge in our classification experiments.

\begin{table}[b]
\vspace{-5mm}
\centering
\begin{tabular}{l c c c } 
 \toprule
 Method & Re-train & Accuracy & NLL \\
 \midrule
 Gal et al.~\cite{Gal2015MCDO} & Yes & 93.2 & 4.79 \\ 
 Gast et al.~\cite{Gast:2018:LPD} & Yes & \rebutall{93.7} & 15.2  \\
 Ours & No & \textbf{94.0} & \textbf{2.65} \\
 \bottomrule \\
\end{tabular}
\caption{\small Benchmark comparison against state-of-the-art on the task of object recognition.}
\vspace{-7mm}
\label{tab:cifar_results}
\end{table}

\noindent\textbf{Closed-Loop Control of a Quadrotor:} In this last experiment, we demonstrate that our framework can be used to fully integrate a deep learning algorithm into a robotics system.
In order to do so, we consider the task of real-time, closed-loop control of a simulated quadrotor.
%
For this task, we deploy a multi-layer perceptron (MLP) to learn compensation terms of a forward-dynamics quadrotor model.
These compensations generally capture model inaccuracies, due to, e.g., rotor or fuselage drag~\cite{faessler2018differential}.

We define the common model of a quadrotor, e.g. the one of Mellinger et al.~\cite{mellinger2011minimum}, as the \emph{nominal} model.
As proposed by previous work~\cite{faessler2018differential}, we add to the linear and angular acceleration of the nominal model $\dot{p}, \dot{\omega}$ two compensations $e_{lin}, e_{ang}$.  
%
However, while previous work~\cite{faessler2018differential} predicts $e_{lin}, e_{ang}$ as a linear function of the platform linear and angular speed $v, \omega$, we propose to predict them as a function of the entire system state $s=(v,\dot{v},\omega,\dot{\omega})$ via an MLP.
Except for the modification of the function approximator (MLP instead of linear), we keep the training and testing methodology of Faessler et al.~\cite{faessler2018differential} unchanged.

We collect annotated training data to predict $e_{lin}, e_{ang}$ in simulation~\cite{Furrer2016}.
Similarly to~\cite{faessler2018differential}, we use a set of circular and lemniscate trajectories at different speeds to generate this data.
%
%
The annotated data is then used to train our MLP, which takes as input $s$ and outputs a 6 dimensional vector $e_{lin}, e_{ang}$.

We use our framework to compute the MLP's prediction uncertainty.
At each time-step, if the uncertainty is larger than a user-defined threshold, the compensations will not be applied.
This stops the network to compensate when uncertain, avoiding platform instabilities.
Specifically, we set this threshold as five times the mean prediction uncertainty in an held-out testing set.

We perform closed-loop experiments on two types of trajectories: a horizontal circular with 4mt radius and an horizontal lemniscate (see Fig.~\ref{fig:closed_loop}).
Both maneuvers, not observed at training time, were performed with a maximum speed of $7\nicefrac{m}{s}$.
%
%
Following previous work~\cite{faessler2018differential}, we use the RMSE metric between the reference and actual trajectory for quantitative analysis.
The results of this evaluation are presented in Table~\ref{table:closed_loop}.
Obviously, the high maneuvers' speed introduces significant drag effects, limiting the accuracy of the nominal model.
Adding a linear compensation model, as in Faessler et al.~\cite{faessler2018differential}, improves performance on the simple circular maneuver, but fails to generalize to the more involved lemniscate trajectory.
Substituting the linear with a non-linear compensation model (MLP) improves generalization and boosts performance in the latter maneuver.
However, applying the compensation only if the network is certain about its predictions (MLP with $\totvar$) additionally increases tracking performance by $2\%$ and $8\%$ on the circular and lemniscate trajectories, respectively.
Indeed, these maneuvers, unobserved at training time, contain states for which compensation is highly uncertain.
Finally, Fig.~\ref{fig:closed_loop} shows a qualitative comparison on the tracking performance of the different methods.
Thanks to the non-linearities and the uncertainty estimation, our approach appears to be closer to the reference trajectory, hence minimizing tracking errors.

\begin{table}[h]
\centering
\begin{tabular}{l c c } 
 \toprule
 Method & Circular & Lemniscate  \\
 \midrule
 Nominal model & 0.271 & 0.299 \\ 
 Faessler et al.~\cite{faessler2018differential} & 0.086 & 0.298\\
 MLP (Ours) & 0.086 & 0.255\\
 MLP with $\totvar$ (Ours) & \textbf{0.084} & \textbf{0.234}\\
 \bottomrule \\
\end{tabular}
\caption{\small Quantitative comparison on trajectory tracking, closed-loop experiments.}
\label{table:closed_loop}
\vspace{-5mm}
\end{table}

\subsection{Practical Considerations}\label{sec:practical}

\begin{figure}
    \centering
    \includegraphics[width=0.95\linewidth]{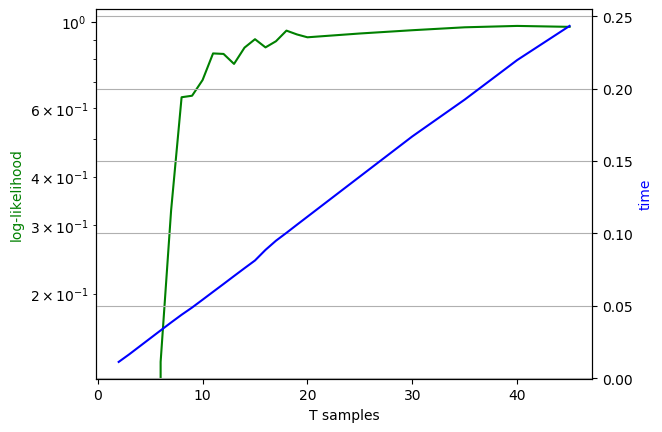}
    \caption{As typical for sampling based approaches, a higher number of samples improves uncertainty estimates. In order to obtain real-time estimates, it is necessary to trade-off performance for speed. For example, in the task of end-to-end steering angle prediction, $T=20$ samples is enough to have good uncertainty estimates and $\approx 10$Hz inference rate.}
    \vspace{-4mm}
    \label{fig:time_vs_likelihood}
\end{figure}

\textbf{Run-time Analysis:} We perform a run-time analysis of our framework to study the trade-off between inference time and estimation accuracy.
Similar to all methods based on Monte-Carlo sampling, our approach requires multiple forward passes of each image to estimate uncertainty.
The larger the number of samples, the better the estimates~\cite{Gal2015MCDO}.
%
%
As it can be observed in Fig.~\ref{fig:time_vs_likelihood}, the quality of variance estimation, measured in terms of NLL, plateaus for $T \geq N$ samples.
In our experiments, we selected $T=20$ as it allows processing at $\approx 10 Hz$, which is acceptable for closed-loop control~\cite{palossi201964mw, kaufmann2018deep}.
%

\rebutall{\textbf{Feed-forward vs Recurrent Models:} Since our derivations are agnostic to the architecture, the proposed framework can be applied to both feed-forward and recurrent models.  
However, while recurrent models can improve performance on sequential tasks, their computational cost for extracting uncertainty is significantly higher: for each Monte-Carlo sample, the entire temporal sequence needs to be re-processed. }

\section{Conclusion}\label{sec:conclusion}

In this work, we present a general framework for uncertainty estimation of neural network predictions.
Our framework is general in the sense that it is agnostic to the  architecture, the learning procedure, and the training task.
%
%
Inspired by Bayesian inference, we mathematically show that our approach tightly couples the sources of prediction uncertainty.
To demonstrate the flexibility of our approach, we test it on several control and vision tasks. 
On each task we outperform state-of-the-art methods for uncertainty estimation, without compromising prediction accuracy.

Similarly to all sampling-based methods~\cite{Gal2015MCDO, Kendall2017}, the main limitation of our approach is that, in order to generate $\totvar$, we need several network forward passes for each input.
\rebutall{This is particularly problematic for recurrent models, which need to unroll the entire temporal sequence for each sample.}
Although we show that this does not hinder real time performance (see Fig.~\ref{fig:time_vs_likelihood}), it still represents the main bottleneck of our framework.
We believe that finding alternative solutions to compute model uncertainty, using, e.g., information theory~\cite{achille2019information}, is a very interesting venue for future work.

{\small
\bibliographystyle{IEEEtran}
\bibliography{bibliography/references.bib}
}


\section{Supplementary Material}\label{sec:supplementary}
\beginsupplement

\subsection{Proof of Lemma III.2}
Consider the probabilistic model of the ADF network and the probabilistic distribution over the input:

\begin{equation}
    \begin{aligned}
    p(\y,\z|\x,\bo) &= p(\y|\z,\bo)p(\z|\x) \\
    p(\z|\x) &\sim \N \left( \z; \x, \sensnoise^{(0)}_{t} \right)
    \end{aligned}
\end{equation}

Let's now place a posterior distribution $p\left(\bo|\mathbf{X}, \mathbf{Y}\right)$ over network weights given the training data $\mathbf{D} = \{\mathbf{X}, \mathbf{Y}\}$. Consequently, the full posterior distribution of the Bayesian ADF network can be parametrized as

\begin{small}
\begin{equation} \label{eq:full_posterior_adf}
    \begin{aligned}
    p\left(\y,\z|\x,\mathbf{X}, \mathbf{Y}\right) &= \left(\int p(\y|\z,\bo) \cdot p(\bo|\mathbf{X}, \mathbf{Y}) d\bo \right) \cdot p(\z|\x) \\
    &= \int p(\y,\z|\x,\bo) \cdot p(\bo|\mathbf{X}, \mathbf{Y}) d\bo
    \end{aligned}
\end{equation}
\end{small}

\noindent
where $p(\y,\z|\x,\bo) = p(\y|\z,\bo) \cdot p(\z|\x) \sim \N (\widehat{\y}_{\bo},\sensnoise^{(l)}_{t} \I_D)$ for each model weights realization $\bo$.
Also, we approximate the intractable posterior over network weights as 
\begin{equation} \label{eq:weights_post}
    p(\bo|\mathbf{X}, \mathbf{Y}) \approx q(\bo) = \text{Bern}(\sBb_1) \cdots \text{Bern}(\sBb_L)
\end{equation}
where $\text{Bern}(\sBb_i)$ is a Bernoullian distribution over the activation of the i-th layer.
Thus,

\begin{small}
\begin{equation} \label{eq:cont_full_posterior_adf}
    \begin{aligned}
    p\left(\y,\z|\x,\mathbf{X}, \mathbf{Y}\right) &\approx \int p(\y,\z|\x,\bo) \cdot q(\bo) d\bo = q(\y,\z|\x)
    \end{aligned}
\end{equation}
\end{small}

We will now prove that our framework actually recovers the total variance by plugging multiple stochastic forward passes with MC dropout in Lemma III.2. \\

\noindent\begin{proof} \label{proof:total_var}
\begin{small}
\begin{alignat*}{2}
&\mathbb{E}_{q(\y,\z | \x)} &&\big( \y\y^T \big) \\
&\myeqone \int \bigg( &&\int \y\y^T \cdot p(\y,\z | \x, \bo) \td \y \bigg) q(\bo) \td \bo \\
&\myeqtwo \int \bigg( &&\Cov_{p(\y,\z | \x, \bo)}(\y_{\bo}) \\
& && + \mathbb{E}_{p(\y,\z | \x, \bo)} (\y_{\bo}) \mathbb{E}_{p(\y,\z | \x, \bo)} (\y_{\bo})^T \bigg) \cdot q(\bo) \td \bo \\ 
&\myeqthree \int \bigg( &&\Cov_{p(\y,\z | \x, \bo)}(\y_{\bo}) + \mathbb{E}_{p(\y,\z | \x, \bo)} (\y_{\bo}) \mathbb{E}_{p(\y,\z | \x, \bo)} (\y_{\bo})^T \bigg) \\
& &&\cdot \text{Bern}(\sBb_1) \cdots \text{Bern}(\sBb_L) \td \sBb_1 \cdots \td \sBb_L \\
&\myeqfour \int \bigg( &&\sensnoise^{(l)}_{t}  \I_D + \widehat{\y}(\x, \sBb_1, ..., \sBb_L) \widehat{\y}(\x, \sBb_1, ..., \sBb_L)^T \bigg) \\ 
& &&\cdot \text{Bern}(\sBb_1) \cdots \text{Bern}(\sBb_L) \td \sBb_1 \cdots \td \sBb_L \\
&\myapproxfive \frac{1}{T} \sum_{t=1}^T &&\sensnoise^{(l)}_{t} + \frac{1}{T} \sum_{t=1}^T \widehat{\y}(\x, \widehat{\sBb}_{1,t}, ..., \widehat{\sBb}_{L,t}) \widehat{\y}(\x, \widehat{\sBb}_{1,t}, ..., \widehat{\sBb}_{L,t})^T
\end{alignat*}
\end{small}

\noindent(1) follows by the definition of expected value. \\
(2) follows by the definition of covariance:
\begin{equation*}
    \text{Cov}(\y) = \EX(\y\y^T) - \EX(\y)\EX(\y)^T
\end{equation*}
(3) follows from Equation~\ref{eq:weights_post}. \\
(4) since $p(\y, \z | \x, \bo) = \N \big( \y; \widehat{\y}(\x, \sBb_1, ..., \sBb_L), \sensnoise^{(l)}_{t} \I_D \big)$. \\
(5) approximation by Monte Carlo integration.\\

\noindent Consequently, from the result just obtained and by the definition of variance, it can be easily shown that the total variance can be computed as:


\begin{small}
\begin{align*}
    \text{Var}_{q(\y|\x)}(\y) &= \EX_{q(\y,\z|\x)} \big( \y\y^T \big) - \EX_{q(\y,\z|\x)}(\y)\EX_{q(\y,\z|\x)}(\y)^T \\
    &\approx \frac{1}{T} \sum_{t=1}^T \widehat{\y}(\x, \widehat{\sBb}_{1,t}, ..., \widehat{\sBb}_{L,t}) \widehat{\y}(\x, \widehat{\sBb}_{1,t}, ..., \widehat{\sBb}_{L,t})^T \notag \\
    &- \EX_{q(\y,\z|\x)}(\y)\EX_{q(\y,\z|\x)}(\y)^T + \frac{1}{T} \sum_{t=1}^T \sensnoise^{(l)}_{t}  \\ 
    &= \frac{1}{T} \sum_{t=1}^T \sensnoise^{(l)}_{t} + (\adfmean_t^{(l)} - \bar{\adfmean})^2 \notag
\end{align*}
\end{small}

The total variance of a network output $\y$ for an input sample $\x$ corrupted by noise $\sensnoise^{(0)}$ is:
\begin{equation}
\totvar = \text{Var}_{p(\y|\x)}(\y)= \frac{1}{T} \sum_{t=1}^T \sensnoise^{(l)}_{t} + (\adfmean_t^{(l)} - \bar{\adfmean})^2
\end{equation}
%

\noindent
which indeed amounts to the sum of the sample variance of T MC samples \textit{(model uncertainty)} and the average of the corresponding \textit{data variances} $\sensnoise^{(l)}_{t}$ returned by the ADF network. 
\end{proof}


%
In conclusion, the final output of our framework is $[\y^*, \mathbf{\sigma}_{tot}]$, where $\y^*$ is the mean of the mean predictions $\{\hat{\y}_t\}_{t=1}^T$ collected over T stochastic forward passes.

\section{Training Details}

\subsection{Implementation} 

\rebutall{Our framework for uncertainty estimation is implemented in Pytorch, and included in the supplementary material.
It will be publicly released upon acceptance.
For training and testing, we use a desktop computer equipped with an NVIDIA-RTX 2080.}
\rebutall{
\subsection{End-to-End Steering Angle Prediction}
The NN architecture used for the task \textit{End-to-End Steering Angle Prediction} is a shallow ResNet that takes inspiration from the DroNet architecture by Loquercio et al.~\cite{Loquercio_2018}.
The network was trained on the Udacity dataset~\cite{Loquercio_2018}, containing approximately $70,000$ images captured from a car and distributed over $6$ different experiment settings, $5$ for training and $1$ for testing. 
A validation set is held out from the data of the first experiment.
For every experiment, time-stamped images are stored from 3 cameras (left, central, right) with the associated data: IMU, GPS, gear, brake, throttle, steering angles and speed. 
For our purpose, only images from the forward-looking camera and their associated steering angles are used. 
The network is trained for $100$ epochs with Adam and an initial learning rate of $1e-3$. 
The loss used for training is an L2-loss, which is also computed on the validation set at every epoch to select the best model. 
The total training time on our aforementioned hardware amounts to $6$ hours .
\subsection{Object Future Motion Prediction}
For the task \textit{Object Future Motion Prediction} we employ a FlowNet2S architecture to predict the future optical flow. Given the frames at time $t-1$ and $t$, it predicts the future optical flow between frame $t$ and $t+1$.
We use the publicly available weights pre-trained on FlyingChairs and FlyingThings3D~\cite{mayer2016large} datasets for the task of optical flow regression to initialize our model, which is then specifically trained for our task for $2000$ epochs (around $15$ hours on our hardware) on the DAVIS 2016 dataset~\cite{Perazzi_2016}.
We used Adam with a learning rate of $1e-3$ and the loss used is the multi-scale L1 metrics.
This multi-scale L1 loss is computed only on the moving object, identified by the segmentation mask. 
As input to the network, we pass the frames at time $t-1$ and $t$, stacked together with the segmentation mask corresponding to frame $t$. 
The optical flow between image $t$ and $t+1$ is used as ground-truth. 
Since DAVIS dataset does not provide optical flow annotations, we use the state-of-the-art FlowNet2~\cite{ilg2017flownet} architecture to collect optical flow annotations for this dataset. 
Instead, the segmentation masks used as auxiliary input are already provided along with the DAVIS dataset. 
At test time, to prove the efficacy of the proposed method, the object mask $M_{t}$ is generated by the state-of-the-art object detector AGS~\cite{wang2019learning}. 
This indeed provides a good indicator of the network performance `in the wild', where no ground-truth object mask is available.}
%

%
\rebutall{
\subsection{Model Error Compensation}
The Multi Layer Perceptron (MLP) used for model error compensation in the task \textit{Closed-Loop Control of a Quadrotor} was trained for 100 epochs on a dataset consisting of data collected by a drone. 
We used Adam with $1e-4$ as learning rate, together with an L1 loss. Training took approximately 2 hours on our hardware.
As input we used $24$ features, each of them being collected at the same time step.  
These features are quaternion odometry, linear velocity odometry, angular velocity odometry, and thrusts. 
The network was trained to learn the linear and angular model error.
The data were collected from three different kind of trajectories: circular, lemniscate and random. 
The circular and lemniscate trajectories were generated with a fixed radius of $4 m$ and different velocities. 
Eventually, the training dataset is composed of almost one million datapoints, consisting of six circular trajectories, six lemniscate trajectories and random trajectories, generated interpolating randomly generated points in the space. 
Each of these trajectories is generated with a fixed velocity per trajectory ranging from $1$ to $8 m/s$.
One tenth of the data was held out for testing and parameter tuning.
}
\rebutall{
\section{Sensitivity to Sensor Noise Estimates}
One of our framework's input consists of the sensor noise variance $\sensnoise^{(0)}$, which is propagated through the CNN to recover data uncertainty on output predictions.
The value of $\sensnoise^{(0)}$ is usually available from the sensor data sheet, or estimated via system identification. In this section, we study the sensitivity of our framework to the precision of the  $\sensnoise^{(0)}$ estimates.
In order to do so, we perform a controlled experiment for the task of steering angle prediction, where each image is corrupted by known Gaussian noise $\N \left(0; \sensnoise^{(0)} \right)$.
In this experiment, our framework has an estimate of the noise variance $\sensnoisehat^{(0)}$, which does not necessarily coincides with the real $\sensnoise^{(0)}$. Specifically, we keep $\sensnoisehat^{(0)}=0.01$, while we let $\sensnoise^{(0)}$ change.
The results of this evaluation are reported in Table~\ref{tab:sens_analysis}.
Perhaps unsurprisingly, performance peaks when the assumed sensor noise coincides with the real input noise magnitude. However, as the difference between the real and assumed noise variance increases, performance gracefully drops for our approach, indicating the robustness of our method to wrong sensor noise estimates. 
Interestingly, Table~\ref{tab:sens_analysis} also shows that our approach deals better than the baselines to increasing magnitudes of the noise. This is due to the coupling of data and model uncertainty enforced by our framework.
}
\begin{table}[t]
\centering
\begin{tabular}{l c c c } 
 \toprule
 Real Sensor Noise $\sensnoise^{(0)}$ & 0.01 & 0.05 & 0.1 \\
 \midrule
 Gal et al.~\cite{Gal2015MCDO} & 0.67 & 0.35 & 0.03 \\ 
 Gast et al.~\cite{Gast:2018:LPD} & 0.74 & 0.37 & 0.04 \\
 Kendall et al.~\cite{Kendall2017} & \textbf{0.99} & 0.3 & -0.11 \\
 Ours ($\sensnoisehat^{(0)}=0.01$) & 0.95 & \textbf{0.41} & \textbf{0.12} \\
 \bottomrule \\
\end{tabular}
\caption{\small Log-likelihood (LL) score for increasing sensor noise. Higher is better.}
\vspace{-4mm}
\label{tab:sens_analysis}
\end{table}
 
\end{document}